\documentclass{article}
\usepackage{arxiv}
\usepackage[utf8]{inputenc}
\usepackage[T1]{fontenc}    
\usepackage[round]{natbib}
\usepackage{enumitem}
\usepackage{graphicx}
\usepackage{multirow}
\usepackage{todonotes}
\usepackage{stmaryrd}
\usepackage{ifthen}
\usepackage{xspace}
\usepackage{amsmath}
\usepackage{amssymb}
\usepackage{amsthm}

\usepackage{xcolor}

\graphicspath{{Pics/}}

\title{Annotation Uncertainty in the Context of Grammatical Change}
\author{Marie-Luis Merten\\
University of Zurich\\ Zurich, Switzerland\\
mlmerten@ds.uzh.ch\\
\And
Marcel Wever\\
Paderborn University\\ Paderborn, Germany\\
marcel.wever@upb.de\\
\And 
Michaela Geierhos\\ 
Universit{\"a}t der Bundeswehr M{\"u}nchen\\ Munich, Germany\\
michaela.geierhos@unibw.de\\
\And
Doris Tophinke\\
Paderborn University\\
Paderborn, Germany\\
doris.tophinke@upb.de\\
\And
Eyke H{\"u}llermeier\\
LMU Munich\\
Munich, Germany\\
eyke@ifi.lmu.de}

\begin{document}

\maketitle

\begin{abstract}
This paper elaborates on the notion of uncertainty in the context of annotation in large text corpora, specifically focusing on (but not limited to) historical languages. Such uncertainty might be due to inherent properties of the language, for example, linguistic ambiguity and overlapping categories of linguistic description, but could also be caused by lacking annotation expertise. By examining annotation uncertainty in more detail, we identify the sources and deepen our understanding of the nature and different types of uncertainty encountered in daily annotation practice. Moreover, some practical implications of our theoretical findings are also discussed. Last but not least, this article can be seen as an attempt to reconcile the perspectives of the main scientific disciplines involved in corpus projects, linguistics and computer science, to develop a unified view and to highlight the potential synergies between these disciplines.    
\end{abstract}

\section{Introduction}
Dealing with language always involves cases of uncertainty, imprecision, and ambiguity.
To a certain extent, one may even say that these aspects constitute prerequisites for an efficient and ``successful'' communication. They are both driving moments in terms of the transformation of language as well as resulting characteristics \citep{byb11,tratro10}. This article addresses uncertainties that are encountered in the context of annotation projects, with a specific focus on historical languages. More precisely, it deals with cases of linguistic ambiguity and overlapping categories of linguistic description -- grouped into tag sets --, but also with the uncertainty on the part of the annotator when tagging (historical) linguistic data.

To reliably reconstruct language change, large historical corpora are indispensable. Only in this way, well-founded statements about the dynamics and distribution of change become possible. Such large corpora projects must -- in avoidance of becoming lifetime projects -- necessarily rely on machine support for annotation. Machine support is also applied in the project that provides the basis for the investigations in this paper. This project aims to reconstruct the language development of Middle Low German (MLG) \citep{tophinke2012,mer18}. To this end, we investigate a suitable large corpus consisting of written texts. A sufficiently large training corpus containing high-quality manually tagged data is needed to train a machine tagger and therefore constitutes a key prerequisite for reliable machine annotation. Here, the idea is to leverage machine learning methodology for constructing an automatic tagger in a data-driven way. By supporting or even replacing a human tagger, machine tagging is supposed to facilitate text annotation and accomplish scalability. 

In the underlying project, annotations are done at the part-of-speech (POS) level as well as at the level of complex form-meaning pairs. Uncertainties are often encountered during this pre-processing step: Text passages are ambiguous and can be tagged in more than one way. Such ambiguities might be inherent in the data and suggest, for example, a context of category change. However, they can also be epistemic, being caused by a lack of knowledge on the side of the (human) annotator. Often, for example, a text is difficult to comprehend, because the annotator is lacking contemporary knowledge. Whatever the source might be, uncertainties of any kind should already be taken into account in the labeling process. Otherwise, training data will be erroneous and possibly mislead the learning algorithm, which eventually creates a faulty machine annotator. In the absence of any indication of uncertainty, a human annotation will be taken as ``ground truth'' by the learner, regardless of whether it is true or false. Therefore, it is required to explicitly inform about the uncertainty, for example by providing a set of plausible candidate tags instead of committing to a single one. Then, the machine itself can decide which of the alternatives is most likely correct, also in light of the other data. Roughly speaking, we take the stance that ``weak'' but correct supervision is better than precise but incorrect supervision.  

In principle, uncertainties can not only be found in historical but also in contemporary language (usage), and this across many languages. Accordingly, the considerations in this article are also relevant for corpus projects and linguistic studies that pursue other concerns and that investigate other languages -- whether in a synchronous or diachronous way. Yet, dealing with historical material or with data of an unknown language can be more challenging due to its alterity. 

This paper aims to examine annotation uncertainty in more detail, hoping to gain insight into the sources and to deepen our understanding of the nature and different types of uncertainty that might be relevant. We are convinced that a sound understanding of annotation uncertainty is an important prerequisite for the development of suitable annotation tools. Therefore, even if practical aspects of software development are beyond the scope of this paper, some practical implications of our theoretical findings will also be discussed. We are also convinced that appropriate handling of annotation uncertainty requires an interdisciplinary approach. Therefore, this article also attempts to reconcile the perspectives of the main scientific disciplines involved in corpus projects: linguistics on the one side, and mathematics and computer science on the other. In fact, even if the modeling of uncertainty has been studied in mathematics and computer science for a long time, and many formalisms have been established, it is not immediately clear to what extent such formalisms are suitable to capture the phenomenon of uncertainty as perceived by linguists in annotation practice.      

The paper is structured as follows. We start with a discussion of the current annotation practice and its limitations in Section~\ref{sec:current}. 
In the next two sections, we provide an overview of how the phenomenon of uncertainty is perceived and dealt with in historical corpus linguistics (Section~\ref{sec:linguistics-view}) and mathematics (Section~\ref{sec:mathematical-view}).  
In Section~\ref{sec:unified}, we propose a unified view and highlight the potential synergies between the disciplines. Section~\ref{sec:practical-implications} is devoted to a discussion of the practical implications of our theoretical considerations.
Section~\ref{sec:conclusion} concludes the paper with a summary and an outlook on future research directions. 

\section{Current annotation practice and limitations}\label{sec:current}

The annotation of corpus data usually foresees that a POS tag is assigned to each token in the corpus \citep[p. 50-54]{kueblerZinsmeister2015}. Either this is done (semi-)automatically or -- and this is especially the case in projects with historical data that cannot (yet) access automated tagger programs at the beginning of their runtime -- the human annotator must accomplish this task.\footnote{Processes of (automated) tokenization and lemmatization, which are also accompanied by uncertainties, will not be discussed here. The focus is on (manual) POS annotation.} Predefined tag sets (e.g., the English Penn Treebank tagset or the Stuttgart-Tübingen tagset for German), which vary concerning their granularity, are often used in this context. The advantage of such standardized tag sets lies in the fact that the POS annotations are comparable across many different projects. Online available automated taggers (e.g., TreeTagger) also use these tagsets. The disadvantage may be that the tags do not always fit the data. Especially in historical projects, this is often the case, where special tag sets tailored to the historical language level are available (for example the Historical Penn Treebank Part-of-Speech tagset or HiTS (historical tagset for German, \citet{dipdonkle13}). In projects with syntactic questions or with Construction Grammar concerns, the annotation process also consists of an annotation step in which syntactic tags or construction tags are assigned to larger syntagms. These tag sets are often developed in the course of the respective annotation project. In general, these research projects are explorative, and no predefined tag sets are available yet. The most comprehensive and high-quality annotated corpus possible is a necessary basis for a subsequent analysis process.

However, an appropriate annotation does not necessarily consist of assigning exactly one POS tag to each token, or exactly one syntactic or construction tag to each syntagma of interest.
For example, we often encounter ambiguous text passages that can result from the transformation of language -- and this in language data of all ages, i.e., in both contemporary and historical corpora.
The (historical) speakers/writers as well as the addressees do not encounter any communication problems due to these ambiguous passages. Ambiguity or polysemy are -- as mentioned at the beginning of this article -- part of everyday language practice.
Contextual information, prior knowledge, knowledge about the type of conversation or the type of text (genre knowledge) as well as further world knowledge, which is referred to for example in the context of conversational implicature \citep{grice3p}, support the understanding. 

Regarding annotation, such ambiguous structures open up various possibilities of (structural) interpretation and thus different annotations.\footnote{Well-known annotation tools like WebAnno \citep{eckart-de-castilho-etal-2016-web} or INCEpTION \citep{tubiblio106270} do not allow annotators to systematically capture  or even quantify the nature of their doubts in the process. I.e., whether it is unclear to which category it can be assigned or whether a category is already provided for this phenomenon or whether the annotator is unsure how to tag it due to lack of experience or co(n)text.} In such contexts, assigning (only) one POS tag to each token with a  high degree of certainty and only one syntactic or construction tag to each syntagma of interest, distorts the linguistic reality. The data annotated in this way neglect a second (or even third etc.) reading or the uncertainty of the annotator, which can be highly relevant for the linguistic analysis process (of grammar change, for example) \citep{hei02}. It should be emphasized that our attention is particularly focused on structural ambiguity, i.e., on uncertainties in the assignment of parts of speech and constructional tagging. It is not about the semantic ambiguity of a word (as a token) that can instantiate different lemmas. 

The following example~(\ref{ex:intro-sample}) from contemporary German demonstrates such cases of structural ambiguity, as they interest us. We face at least two ways of analyzing the grammatical structure (in a first approximation), captured by reading A and B, respectively (here and below the HiTS tag set is used, see \citet{dipdonkle13}):\\[\baselineskip]
\begin{tabular}{r p{9cm}}\label{ex:intro-sample}
     & \textit{Der nachfolgende Abschnitt handelt vom \textbf{in die Stadt fahren}.} \\
     Reading A: & The following section is about \textbf{into the city drive}.\\
     Reading B: & The following section is about \textbf{driving into the city}.\\
\end{tabular}\\

\begin{table}[]
    \centering
    \resizebox{\textwidth}{!}{
    \begin{tabular}{|p{1.8cm}|c|c|c|c|c|c|c|c|c|}
        \hline
         & Der & nachfolgende & Abschnitt & handelt & vom & in & die & Stadt & fahren \\
         \hline
         POS level & DDART & ADJA$<$VVPS & NA & VVFIN & APPR & APPR & DDART & NA & ? \\
         \hline
         Syntagma level& \multicolumn{3}{c|}{NP} & & & \multicolumn{4}{c|}{?} \\
         \hline
    \end{tabular}
    }
    \caption{During manual annotation, each word must be assigned a POS tag (an extended version of the HiTS tag set is used here; \citet{dipdonkle13}) and each larger syntagma of interest a syntactic or construction tag. Ambiguous passages present challenges in this context (marked with a question mark).}
    \label{tab:my_label}
\end{table}

In annotation projects that follow the standard procedure (one tag per token/per syntagma of interest), the token ``\textit{fahren}'' at POS level and the phrase ``\textit{in die Stadt fahren}'' at syntagma level can be challenging and lead to a need for discussion. Whether the phrase ``\textit{in die Stadt fahren}'' should be annotated as a complex noun phrase or still closer to its verbal origin is debatable. It is also not possible to determine whether ``\textit{fahren}'' is to be understood as an infinite verb or already as a (deverbal) noun (cf.\ Table 1). Annotators can be uncertain at this point, and forcing them to provide a determinate annotation, which is not given in this way, may falsify the data. Indeed, the annotation can be an uncertain one, ranging between two categories $A$ and $B$. Both could be `correct' depending on the underlying grammar model and also on the annotation guidelines. And this hits another nail on the head: the majority of project-internal guidelines would most likely -- based on several formal and/or functional criteria -- exclude one of the possibilities: either the older or the newer one (see for exceptions \citep[p.~92f]{dipdonkle13}. on a document-specific vs. lemma-specific annotation). Such a simplifying and erroneously disambiguating annotation thus blocks appropriate access to highly relevant (ambiguous) text passages in a later analysis process. This phenomenon area, where ambiguous text passages occur, is only one example of uncertainty, which, together with other aspects, will be discussed in more detail in the following. Regarding machine learning, it should be pointed out that training data obtained through misleading disambiguation may lead to incorrect generalizations. Text passages, in which category changes similar to those in the above example occur, can only be interpreted by the machine in one way -- it does not know about any second possibility. 

Before proceeding, let us again emphasize that the focus of this contribution is not on the analysis of a specific corpus. Instead, and more fundamentally, it aims at understanding the phenomenon of ``annotation uncertainty'' -- both from a linguistic and a mathematical perspective. In the following, we will reconcile these two perspectives and bring together those disciplines that often cooperate in corpus projects (linguistics and computer science). The underlying corpus of Middle Low German texts has a more illustrative character. It is used to substantiate concrete problem cases that are related to the phenomenon of uncertainty in the field of corpus annotation.

\section{Uncertainty in Historical (Corpus) Linguistics}\label{sec:linguistics-view}

\subsection{Project Context and Underlying Corpus}
The project that motivates this paper investigates the language development of Middle Low German (MLG) -- a written language fully developed at the beginning of the Early Modern Period (16$^{th}$ century). As in the case of High German, the road to this status as a supra-regional written language was long. It can be traced employing a sufficiently large quantity of text. To this end, a correspondingly extensive corpus of texts from the 13$^{th}$ to 17$^{th}$ centuries is  examined (1.4 million tokens). The focus is on phenomena of grammatical change: written language structures emerge, so-called (complex) constructions, as focused by (radical) construction grammar \citep{cro01}. Examples are subordinate clause constructions, various prepositional techniques, the nominal phrase condensing procedures, and conjunctional adverb constructions. Constructions are always based on the coupling of form and function/meaning -- both levels must, therefore, be taken into account in corpus analysis. Historical legal texts, especially codifications of municipal law, are particularly instructive when investigating language development \citep{maas2010}. Therefore, MLG legal texts make up the bulk of the corpus. Besides, the corpus consists of MLG pharmacopoeias, which are the result of a further professional(izing) practice in which literate structures emerge. The first Early New High German legal texts arising in the Low German language area are also included in the corpus (135,000 tokens). Based on these texts, the thesis is investigated whether the legal writers, after the written language shift (beginning in the 16$^{th}$ century), continue to use the MLG grammar they are familiar with and only realize lexemes in Early New High German. For a more detailed description of the project aims and the reasons behind the choice of corpus see \citep{seemann-etal-2017-annotation}.

\subsection{Annotation Uncertainties}\label{sec:annotationuncertainties}
The underlying corpus is annotated at both word level (i.e., part of speech / POS) and construction level. At this level of constructions we are dealing with more comprehensive syntagms, often recurrent sequences of word forms and word types that are linked to certain (abstract) functions. In particular, the project focuses on conditional constructions in a broad sense, as they are typical for legal texts and pharmacopoeias.  This cluster of constructions is subject to a far-reaching change -- from a more aggregative to an integrative organization -- which manifests itself in the underlying corpus. For annotation on the POS level, an extended version of the HiTS tag set \citep{dipdonkle13} is used. The extensions concern, among other things, the verbal domain (copula verbs) and adjectives derived from present participles (see Table 1). The HiTS tag set claims to be applicable across several historical language levels of German and thus does justice to the alterity of the corpus data -- in comparison to contemporary language material. Our POS annotations can, therefore, be compared with data from the usual reference corpora (such as the reference corpus Middle Low German/Lower Rhenish (1200-1650) or the reference corpus Early New High German (1350-1650)). On the level of constructions, the project employs a tag set tailored to the writing area in focus. This is because the examined form-function pairs can be regarded as legal writing techniques, and therefore are domain-specific.  Moreover, these form-function pairs are in a state of flux, and this too must find its way into the tag set and the annotation guidelines for selecting the corresponding tags. No predefined tag set can be applied at this point. Instead, a tag set specifically customized for the material is used.

In annotation, we often encounter challenging passages resulting from various uncertainty sources, for example, the following text passages. Here we face ambiguous contexts allowing different (structural) interpretations (cf.\ on reanalysis \citet[p. 542f]{trousdale2012}). These contexts can be classified as bridging contexts in the sense of \citet{hei02}: They ``trigger an inferential mechanism to the effect that, rather than the source meaning, there is another meaning, the target meaning, that offers a more plausible interpretation of the utterance concerned'' \citep[p. 84]{hei02}.\\[\baselineskip]
\begin{tabular}{r p{9cm}}
    \multicolumn{1}{l}{(a)} & \textit{We sik eruegu+odes vnderwint · oder an sprikt ·} \textbf{[[\textit{na}]$_\text{APPR}$ [\textit{deme}]$_\text{DDS}$ / [\textit{dat}]$_\text{KOUS}$]$_\text{KOUS}$} \textit{it im vor delet is vor gherichte · Dat is en vredebrake} \citep{goslar} \\
     Translation: & Whoever takes possession of hereditary property or lays claim to it, \textbf{after it that (= after) }it has been deprived of it by judgment in court: this is a breach of the peace\\
\end{tabular}\\[\baselineskip]
\begin{tabular}{r p{9cm}}
    \multicolumn{1}{l}{(b)} & \textit{Sterft en man de en echte wif hinder sek let . de kindere to samende hebben .} \textbf{[[\textit{de}]$_\text{DDART}$ [\textit{wile}]$_\text{NA}$]$_\text{KOUS}$} \textit{dat wif nenen anderen ghaden nymt . ne darf se mit den kinderen nicht delen} \citep{goslar} \\
     Translation: & If a man dies leaving behind a woman, who has children together. \textbf{the while (= as long as)} the woman does not take another husband she may not share with the children \\
\end{tabular}\\[\baselineskip]
\begin{tabular}{r p{9cm}}
    \multicolumn{1}{l}{(c)} & \textbf{[[\textit{De}]$_\text{DDART}$ [\textit{wile}]$_\text{NA}$]$_\text{KOUS}$} \textit{de vrowe nenne anderen gaden ne nimt . So is se irer kindere vormu(n)de . icht se to gosler wont} \citep{goslar} \\
     Translation: & \textbf{The while (= as long as)} the wife does not take another husband. So she is the guardian of her children if she lives in Goslar. \\
\end{tabular}\\


The highlighted units are recurrently used in the investigated texts. They make up the lexical anchors of the varying constructions. What is interesting now is their categorization. These units are (structural and therefore also semantically in a broad understanding) ambiguous: On the one hand, the tokens \textit{na} + \textit{deme} + \textit{dat} and \textit{de} + \textit{wile} (+ facultative \textit{dat}) can be analyzed as three/two individual lexemes (\textit{na deme dat}: preposition + determiner + primary subjunction; \textit{de wile} (\textit{dat}): determiner + noun (+ facultative primary subjunction)) on the POS level. On the other hand, these combinations are relatively fixed chunks in the sense of \citet{byb10}: ``Once word sequences such as \textit{be going to} or \textit{in spite o}f have become frequent enough to be accessed from cognitive storage and produced as units, they begin to become autonomous from the words or morphemes that compose them'' \citep[p.~84]{byb11}. These function word groups already exhibit their future function as a (temporal/causal) (secondary) subjunction. At the beginning of their genesis (and chunking), the function word groups \textit{na deme dat} and \textit{de wile (dat)} construe a temporal relationship between two units, which can be understood as sentences -- a typical characteristic of subjunctions. In the later period, they are used increasingly abstractly, i.e., as causal relators, their form also becomes shorter as processes of formal erosion take place. 

\subsubsection{Overlapping categories and the gradualness of change}
What do these examples (a to c) tell us about sources of uncertainty in the (corpus) linguistic context? These examples provide evidence for a special type of non-classifiable cases. The realizations of certain structures can be categorized as in-between phenomena. The corresponding instances belong to more than one radial category. That means, these realizations exhibit properties of at least two categories~A (e.g.,\ noun phrase (\textit{de wile}) or prepositional structure (\textit{na deme dat})) and~B (e.g.,\ secondary subjunction) and can be located in the overlapping area of them. These properties can be both formal and functional semantic since grammatical categories (word types, constructions, and so on) are composed of properties at both levels in a Construction Grammar model of language. 
The boundaries of these categories are blurred and they often change over time \citep[p.~520]{dip15},
for example as a result of constructional change or constructionalization (i.e., new constructions emerge). Both processes have in common that they are gradual in nature, often they merge into each other \citep[p.~32]{trousdale2013gradualness}. Normally, only one constructional feature changes at a time, for example, regarding the formal dimension, the function, the domain specificity, or frequency of a construction. The developments observable in language use are (very) slight. However, with a certain time lag, larger changes based on these smaller steps become visible. As we are analyzing data that spans over 300~years, we are faced with numerous micro-changes and tiny-step transmissions \citep[p.~74f]{tratro13}, but can also trace changes on the level of macro constructions.

Interesting in this context is the synchronous perspective: Within one single text, the different stages of constructional change can ``surface as contextually defined variants'' \citep[p.~409]{heinar10}.
Constructions of different ages exist side by side. However, frequency differences between older and younger variants that belong to one constructional path may become apparent. Also, interesting shifts in the frequency of occurrence over the period under investigation can occur. In principle, the following applies: The distinction between synchrony and diachrony is not sharp, they ``have to be viewed as an integrated whole'' \citep[p.~105]{byb10}. 
At any point in time, there will be units that do not fit squarely into the linguist’s categories, the coexistence of differently aged structures, of better and less suitable category members within a time period/text suggests to model categories as radial and fuzzy. They are gradient in the sense of \citep{aar07}. 

\subsubsection{Types of categorial gradience}
As already outlined, the gradualness of change synchronically manifests itself in small-scale variation and categorial gradience \citep{tratro10, byb11}:
``[A]t any moment in time changing constructions contribute to gradience in the system'' \citep[p.~75]{tratro13}.
Under this premise, (grammatical) categories -- e.g.,\ noun phrase, preposition, subjunction, complex constructions and so on -- typically have a gradient structure; they can be viewed as prototype categories. Characteristic properties are the gradation of typicality, the already mentioned fuzzy boundaries and potential overlaps with other classes \citep{lak87, tay03} -- similar to the mathematical notion of fuzzy sets (cf.\ Section 2.1.3).
In this regard, \citet{aar07} 
differentiates between (a)~intersective and (b)~subsective gradience. Both phenomena are relevant for our project, and likely to be significant in other project contexts, too (including projects with a contemporary language focus).

\begin{figure}[t]
    \centering
    \includegraphics[width=.8\linewidth]{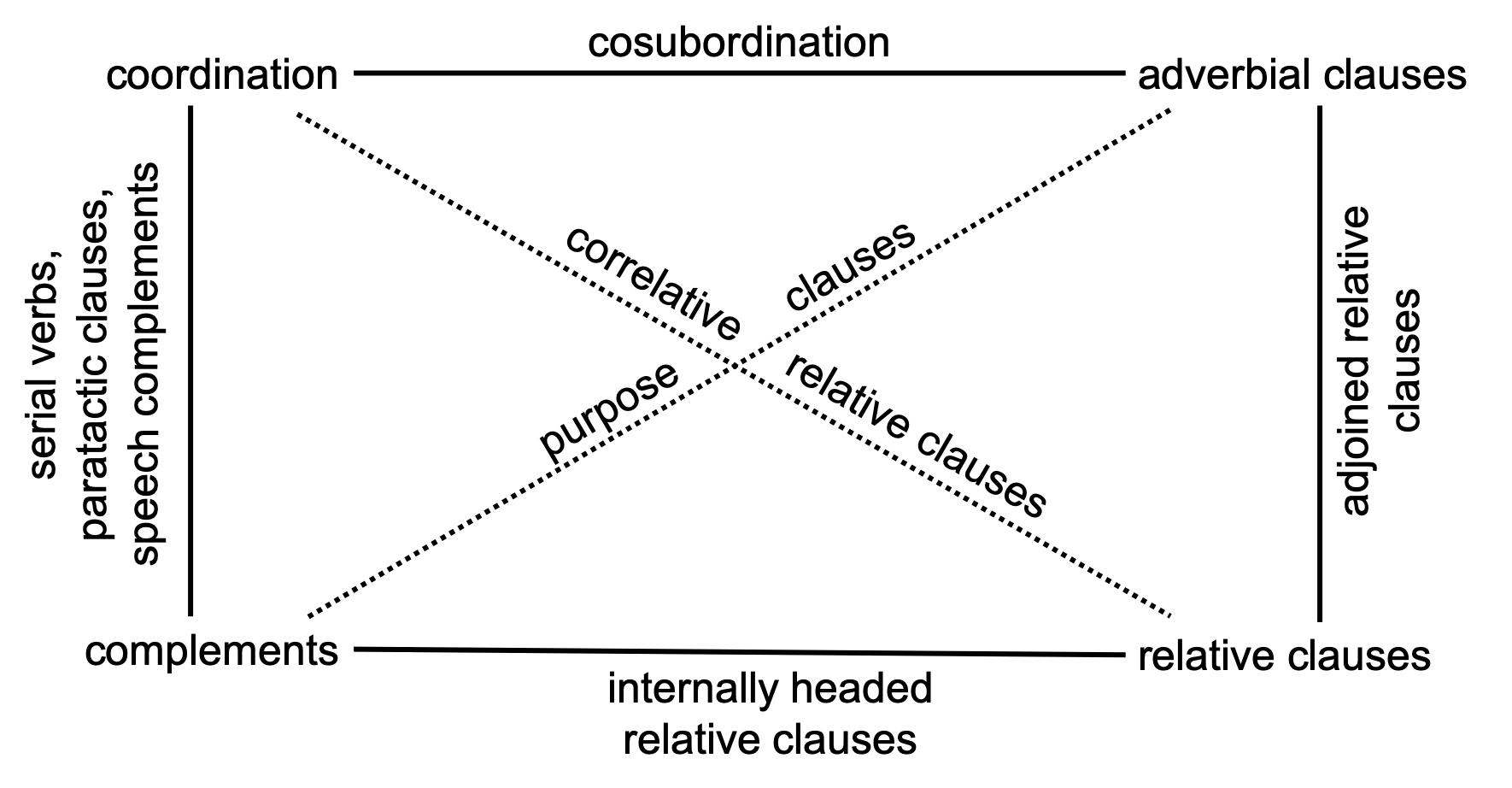}
    \caption{coordination-subordination continuum modeled by Croft (2001: 322)}
    \label{fig:croft}
\end{figure}

Intersective gradience is an inter-categorial phenomenon. It refers to the preference for a non-discrete view of linguistic categories (as `represented' by tags). This view is essentially shaped by the assumption that there are cases of indeterminacy and no sharp distinctions (blurred boundaries) between certain classes (e.g.,\ POS, constructions, and so on). The investigated categories can be seen as grading into or converging on one another (concerning formal and/or functional-semantic aspects). According to \citet[p.~18]{lan87}, 
these convergent categories form a continuous spectrum (or field) of possibilities, whose segregation into distinct blocks is necessarily artefactual. Continua such as those between coordination and subordination or relative and complement clauses – two phenomena we are interested in – illustrate this type of gradience as shown by \citet{cro01} (cf.\ Figure~\ref{fig:croft}).

To demonstrate this type of gradience, we turn to a concrete example from the field of (emerging) complex sentences: Sometimes it cannot be clearly distinguished whether a paratactic or hypotactic sentence organization is present. For example, formal markers such as an introductory word (\textit{wanne} (`if')) and verb-final position in German may indicate subordination, but the corresponding structures are neither correlatively resumed nor topologically integrated into a supposed matrix sentence -- by taking a prefield position, for instance. Instead, they are arranged relatively loosely in a pre-prefield position -- an argument for aggregation. With the following example, we face such a case: The \textit{wanne}-clause A and the subsequent verb-initial conditional clause B together with the verb-second clause C are organized aggregatively:\\[\baselineskip]
\begin{tabular}{r p{9cm}}
    \multicolumn{1}{l}{(d)} & [\textit{Wanne en vser borghere sterft de eyne echte husvrowen let}]$_\text{A}$ $\cdot$\ [\textit{wel de vrowe then in clostere} $\cdot$\ \textit{spettal eder conuent}]$_\text{B}$ $\cdot$\ [\textit{de scal} $\ldots$]$_\text{C}$ (Goslar 1350) \\
     Translation: & If one of our citizens dies, who leaves behind a wife. If the woman wants then to move to a cloister, hospital, or convent, she should $\ldots$ \\
\end{tabular}\\


Intersective gradience (and related ambiguous contexts in language material) complicate the necessary assignment of categories for further corpus analysis. Human annotators have to be aware of different (almost equal) alternatives. 

Subsective gradience refers to categories that consist of a central core and a peripheral (boundary) area. Here, the focus of attention shifts from inter-categorial relations to the intra-categorial perspective: The structure of a single category and the observable categorial shading are brought into focus. As prototype categories, some instantiations of these categories (as specific language material) are ideal examples. Such cases can be seen as core phenomena. Others have to be analyzed as peripheral elements because they deviate more or less from the prototype \citep{tay03}.
These categories exhibit different degrees of category membership, not every member is equally representative of its class. However, these prototype categories also change over time; the prototype preposition in MLG, for example, is to be modeled differently than the prototype of this category in New Low German. Regarding the development, usage, and evaluation of tag sets, this has to be taken into account. Depending on the project aims, there might be scenarios where either a finer-grained tag set or other solutions to annotate peripheral cases are needed \citep{mertentophinke2019}. 

\subsubsection{The human annotator as a source of (subjective) uncertainty}
What else has to be considered in a corpus-based analysis of language, especially historical language? When analyzing historical (written) data, human annotators always act as non-native speakers \citep[p.~86]{dipdonkle13}.
Linguists benefit from their expertise concerning established ways of interpreting and classifying linguistic phenomena. Yet, they may lack the cultural and pragmatic facts obvious to a contemporary reader \citep{den17}.

Additionally, there might be unknown structures presumed as unusual or even ungrammatical when analyzing them for the first time. Instead of classifying such unknown structures as erroneous cases (writing mistakes and so on), we assume human annotators to have (yet/initially) incomplete knowledge (as epistemic uncertainty) and develop more and more expertise when investigating a growing number of historical texts. For instance, it is not until one annotates a sufficiently large number of texts from the 16/17$^{th}$ century that one realizes that the missing finite auxiliary verb in a subordinate clause complies in fact with the typical realization for certain types of dependent clauses (e.g.,\ relative clauses). Actually, its absence can be interpreted as a marker for subordination in this period of time. Here, the criterion of (efficient) frequency is pivotal: A high frequency of occurrences (e.g.,\ realization without a finite auxiliary verb) can mirror the typicality of the respective structure. We perceive recurrent combinations of elements and linguistic characteristics as patterns and store them as common coding techniques. Very likely, a construction is an entrenched entity at the time of its frequent usage (for a further discussion concerning the connection of frequency and entrenchment see \citet{sch10}).

Especially in this context, linguists have to be aware of a comparative fallacy that emerges when researchers fall into the error of investigating one language by comparing it to another, for example, their native language \citep{ble83}. 
Here, the historicity of the investigated language has to be given serious consideration. Particularly, those linguists analyzing historical stages of languages closely related to their mother tongue (e.g., New High/Low German vs. Early New High German and Middle High/Low German) run into the danger of comparing their (modern) language with the historical data. As a consequence, they may only perceive (salient) deviant forms and overlook less salient ones, because they differ only to a lower extent but are typical patterns. Regarding this source of misinterpretation, human annotators have to be sensitized to the problems of such a misleading comparison. Historical languages must be viewed in their common structures/constructions, characteristics, and functionalities.

This linguistic overview is followed by a sketch of the mathematical and computer science perspective on the phenomenon of (annotation) uncertainty. In Section 5, the two perspectives will be combined synergistically.

\section{Mathematical Modeling of Uncertainty}\label{sec:mathematical-view}
The notion of uncertainty has been studied in various branches of science and plays a major role in disciplines like economics, engineering, and the social sciences, typically in the appearance of applied statistics. Both philosophically and mathematically, uncertainty is most closely connected with the notion of \emph{probability}, which has a long scientific history \citep{hack_te}. Yet, with the emergence of more recent fields such as artificial intelligence, it became clear that uncertainty is not restricted to the concept of \emph{chance}. Instead, uncertainty can occur in various guises, such as inconsistency, incompleteness, imprecision, and vagueness, which may call for different formalizations and mathematical calculi \citep{krus_ua}. Correspondingly, the contemporary literature on uncertainty is rather broad.

\subsection{Frame of Discernment and Ground-Truth}

Formal inference and reasoning about uncertainty presume a formal (mathematical) \emph{model} of a real-world phenomenon, and modeling necessarily involves a simplification and abstraction of the system under consideration. A first important decision in this regard concerns an underlying reference set $\Omega$, sometimes called the \emph{frame of discernment} \citep{shaf_am}. This set consists of all possible ``states of the world'' that ought to be distinguished in the current context. The elements $\omega \in \Omega$ are supposed to be exhaustive and mutually exclusive. Moreover, it is assumed that exactly one of them, $\omega^*$, corresponds to the ``ground truth''. For example, $\Omega = \{ \text{head}, \text{tail} \}$ in the case of coin tossing, $\Omega = \{ \text{win}, \text{loss}, \text{tie} \}$ in predicting the outcome of a football match, or $\Omega = \{ \text{noun}, \text{verb} , \ldots \}$ in POS-tagging. For ease of exposition, and to avoid measure-theoretic complications, we will subsequently assume that $\Omega$ is a discrete (finite or countably infinite) set, like in the previous examples. In principle, $\Omega$ could of course also be a continuous set, for example, a numerical scale such as $[0,250]$ used to quantify the speed of a car.  

As simple as the idea of a frame of discernment formalized as a set $\Omega$ may appear, one should be aware that it (implicitly) relies on assumptions that may not necessarily be taken for granted. In particular, this concerns the assumption of the existence and the uniqueness of a single ground truth\,---\,recall, for example, our discussion about categorical gradience in Section \ref{sec:annotationuncertainties}.
Even if this assumption can be assured, it may sometimes imply the need for complex mathematical structures. For example, consider the problem of modeling uncertainty about the languages spoken by a specific person. In this case, each element $\omega$ is already a set by itself, namely a subset of an underlying set of languages $\Omega'$, and $\Omega$ is the power set of $\Omega'$. As an aside, note that the definition of $\Omega'$ might not be straightforward either, as it requires a decision about what counts as a language and what not, whether two dialects should be considered as different languages or not, etc.  As can be seen, depending on what aspects ought to be captured, the frame of discernment can often be modeled in various ways and at various levels of detail. 

Obviously, the level of detail or granularity of $\Omega$ will also have a direct influence on the uncertainty about $\omega^*$. For example, if only a rough distinction is made between outside temperatures above and below $0^\circ$\,C, i.e., $\Omega = \{ \text{cold}, \text{warm} \}$, one might be very sure about the ground truth, whereas the precise temperature on the scale $\Omega = \{ 0, \pm 1^\circ, \pm 2^\circ,  \ldots  \}$ might not be known precisely. The idea of modeling on different levels of abstraction and with various degrees of granularity will be further discussed in Section \ref{sec:rough}.

In addition to the uniqueness of a ground truth, the notion of a frame of discernment also comes with the assumption of exhaustiveness of $\Omega$: the current context is always captured by at least one element $\omega \in \Omega$. In the literature, this is sometimes called the ``closed world assumption'', whereas a possibly non-exhaustive $\Omega$ is considered as an ``open world'' \citep{deng_ge14}. Obviously, the closed world assumption requires the anticipation of all contingencies, which is not always easy to assure. In the case of annotation, for example, $\Omega$ corresponds to the tag set, and while exhaustiveness might still be plausible for POS tagging, it is certainly less so for syntactic constructions. Likewise, in our spoken language example, all languages that might be spoken by a person must be known beforehand.  Although the distinction between closed and open world may look technical at first sight, it has important consequences concerning the representation and processing of uncertain information. For example, while the empty set $\emptyset$ is logically excluded as a valid piece of (uncertain) information under the closed world assumption, it may suggest that the true state $\omega^*$ is outside $\Omega$ under the open world assumption.

\subsection{Uncertainty Measures and Calculi}

In addition to the assumption of a unique (albeit unknown) ground truth $\omega^*$, another characteristic property of formal models of uncertainty is the \emph{quantification} of the uncertainty about $\omega^*$ in terms of an \emph{uncertainty measure}. Most commonly, a measure $\mu$ assigns a value $v \in V$ to each subset $A \subseteq \Omega$, which can be interpreted as a degree of belief or evidence that $\omega^* \in A$. More specifically, given a frame of discernment $\Omega$, each subset $A \subseteq \Omega$ (called an ``event'' in probability theory) can be associated with the logical proposition that the ground truth corresponds to one of the alternatives in $A$, and $v = \mu(A)$ can be considered as a valuation of this proposition. In POS-tagging, for example, $A = \{ \text{verb}, \text{adjective} \}$ would suggest that the word under consideration is either a verb or an adjective. 

Different uncertainty calculi distinguish themselves by mathematical properties of the measure $\mu$, as well as specific choices of the underlying uncertainty scale $V$. For example, classical probability theory quantifies uncertainty in terms of real numbers in the unit interval $V = [0,1]$ and assumes $\mu$ to be additive (i.e., $\mu(A \cup B) = \mu(A) + \mu(B)$ for all $A, B \subseteq \Omega$ such that $A \cap B = \emptyset$). In contrast to this, so-called possibility theory \citep{dubo_pt} works with measures that are ``maxitive'' instead of additive, in the sense that $\mu(A \cup B) = \max\{ \mu(A) , \mu(B) \}$ for all $A, B \subseteq \Omega$ \citep{shil_mm71,dubo_pt06}. Besides, possibility theory assumes a weaker, ordinal scale of uncertainty degrees, for example $V = \{ \bot, u , l, \top \}$, where $\bot$ may stand for impossible, $u$ for unlikely, $l$ for likely, and $\top$ for certainly true. Both calculi, probability and possibility theory, offer different rules for combination of uncertain pieces of information, conditioning, etc.

Different mathematical properties of these (and other) calculi are accompanied by different semantic interpretations, rendering each calculus more or less suitable for modeling certain phenomena of uncertainty. In particular, due to its additive nature, probability measures are well-suited for modeling the phenomenon of \emph{chance} or \emph{randomness}, essentially capturing \emph{aleatoric} uncertainty. However, they are arguably less suitable for representing \emph{epistemic} uncertainty or \emph{ignorance} in the sense of a lack of knowledge \citep{dubo_rp96}. For example, the case of \emph{complete ignorance} is typically modeled in terms of the uniform measure $\mu(A) \equiv |A|/|\Omega|$ in probability theory\footnote{This is justified by the ``principle of indifference'' invoked by Laplace, or by referring to the principle of maximum entropy.}. Then, however, it is not possible to distinguish between (i) precise (probabilistic) knowledge about a random event, such as tossing a fair coin, and (ii) a complete lack of knowledge, for example, due to an incomplete description of the experiment. This was already pointed out by the famous Ronald Fisher, who noted that ``\emph{not knowing the chance of mutually exclusive events and knowing the chance to be equal are two quite different states of knowledge}''. As an illustration, consider the following question: What does the word ``tarakimu'' mean in the Swahili language, heads or tails? The possible answers are the same as in coin flipping, and one might be equally uncertain about which one is correct. Yet, the nature of uncertainty is very different.

Possibility theory appears to be more appropriate in this regard. In particular, complete ignorance is conveniently modeled by the measure $\mu(A) = 1$ for all $A \neq \emptyset$, suggesting that all alternatives are completely plausible. More generally, possibility theory allows for modeling purely set-valued information in the form of a constraint $\omega^* \in C \subseteq \Omega$, which is equivalent to the measure $\mu$ such that $\mu(A)= 1$ if $A \cap C \neq \emptyset$ and $\mu(A) = 0$ otherwise. In a sense, probability and possibility measures can be seen as two extremes on the scale of uncertainty representations, with various other theories of uncertainty in-between, including imprecise probability \citep{wall_sr}, random sets \citep{math_rs,nguy_or78}, and evidence theory \citep{shaf_am,smet_tt94}.


\subsection{Vagueness, Fuzziness, and Graded Notions of Truth}

The uncertainty calculi discussed so far allow for representing incompleteness of information, a lack of knowledge about the sought ground truth, as well as the phenomenon of randomness. Yet, they are still grounded in classical logic and set theory, and hence build on the notion of \emph{truth} as a \emph{bivalent} concept: Logically, a proposition is either true or false, but nothing in-between. This conception, which pervades modern science and thinking, has a longstanding tradition in Western philosophy. 



    \begin{figure}
    \centering
    \includegraphics[width=0.6\linewidth]{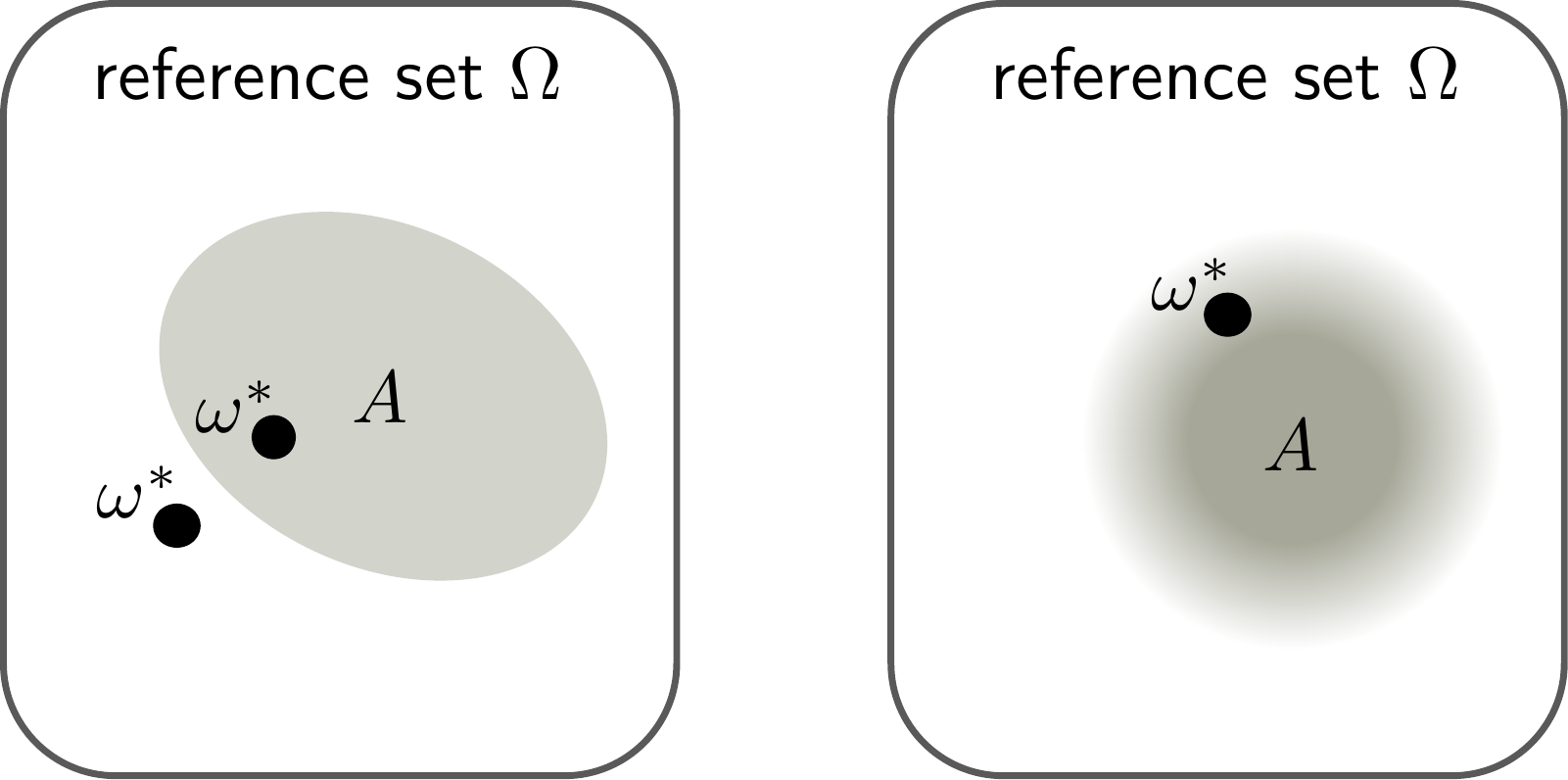}
    \caption{Right: Uncertainty calculi such as probability theory are based on the classical logical notion of bivalent truth. Thus, propositions are associated with subsets $A$ of the reference set $\Omega$, and are either true ($\omega^* \in \Omega$) or false ($\omega^* \not\in \Omega$)\,---\,the uncertainty concerns the ground truth $\omega^*$, which is supposed to be unknown. Right: In fuzzy logic, propositions or events $A$ can be characterized in terms of non-classical sets with soft boundaries. Thus, even if $\omega^*$ is precisely known, the truth of the proposition $\omega^* \in \Omega$ is a matter of degree and cannot be decided unequivocally.}
    \label{fig:compl}
    \end{figure}

While formal systems based on bivalent logic have proved extremely useful in the scientific terrain, where they paved the way for the amazing success of the exact and engineering sciences in the last century, there are many fields of application in which the bivalence of truth can be called into question. In fact, it was already noticed by Bertrand Russell in 1923 that ``All traditional logic habitually assumes that precise symbols are being employed. It is therefore not applicable to this terrestrial life, but only to an imagined celestial existence'' \citep{russell}. Roughly speaking, this is because of the vagueness and ambivalence of the concepts dealt with in these fields: for the intension of these concepts, there is rarely a precise extension (in the sense of a set of real objects belonging to that concept) in the real world. 
For example, what does it mean that a text has a ``positive sentiment'', or that a person speaks a certain language? One may argue that, for a fixed person P and language L, the proposition ``P speaks L'' might be neither completely true nor completely false, but possibly something in-between. Correspondingly, the set of people speaking language L, i.e., the extension of the concept ``L-speaking people'', might be a set with non-sharp boundaries.

The core idea of the mathematical notion of a \emph{fuzzy set}, as introduced by L.A.\ Zadeh in his seminal paper \citep{zade_fs65}, is to allow for partial membership and ``soft'' class boundaries of that kind. Thus, fuzzy sets formalize the idea of graded class membership, according to which an element can partially belong to a set. In conjunction with generalized logical (set-theoretical) operators and derived notions like a fuzzy relation, the concept of a fuzzy set can be developed into a generalized set theory, which in turn provides the basis for generalizing theories in different branches of (pure and applied) mathematics.
The relation between (classical) set theory and fuzzy set theory is analogous to the relation between (classical) bivalent logic and fuzzy logic, which allows for taking truth degrees from richer scales and operates on such scales with generalized logical operators \citep{haje_mf}. An important example of such a scale is the unit interval $[0,1]$, which comprises an infinite number of truth degrees. Here, a degree of 1 would be interpreted as ``completely true'', a degree of 0 as ``completely false'', whereas any value in-between is expressing a certain degree of truth. 


Formally, a fuzzy subset $A$ of a reference set $\Omega$ is identified by a so-called \emph{membership function}, often denoted $\mu_A(\cdot)$, which is a generalization of the characteristic function $\mathbb{I}_A(\cdot)$ of an ordinary set $A \subseteq \Omega$. For each element $\omega \in \Omega$, this function specifies the degree of membership of $\omega$ in the fuzzy set; it can be interpreted as the truth degree of the proposition that $\omega \in A$. As already said, membership degrees $\mu_A(\omega)$ are often taken from the unit interval [0,1], i.e., a membership function is a mapping $\Omega \rightarrow [0,1]$. In principle, however, more general membership scales, including ordinal scales or complete lattices, can be used. One way to combine fuzzy sets with the measure-theoretic approach to uncertainty modeling is to extend the domain of an uncertainty measure $\mu$ from the set of subsets of the frame of discernment $\Omega$ to the set of fuzzy subsets of $\Omega$. For example, if $\mu$ is a probability measure, this gives rise to the notion of \emph{fuzzy probability} \citep{buckley}.  

\subsection{Granularity}\label{sec:rough}

Another extension of classical set theory, so-called rough set theory \citep{pawl_rs82}, is rooted in the idea of a granular representation due to the \emph{indiscernibility} or \emph{indistinguishability} of elements in $\Omega$. Thus, the idea is that, due to limited capabilities in perception or measurement, certain elements $\omega$ and $\omega'$ (states of the world) might not be distinguishable. Mathematically, this is captured in the form of an equivalence relation $I \subseteq \Omega \times \Omega$, where $(\omega , \omega') \in I$ means that $\omega$ and $\omega'$ cannot be distinguished. The relation $I$ gives rise to a partition of $\Omega$ into equivalence classes or ``granules'' $[\omega]_I = \{ \omega' \in \Omega \, | \, (\omega , \omega') \in I \}$. 

In light of the limited resolution imposed by the (indiscernibility) relation $I$, it might not be possible to represent every subset $A \subseteq \Omega$ exactly. Instead, a set $A$ is represented in terms of its lower approximation $\underline{A} = \{ [\omega]_I \, | \, [\omega]_I \subseteq A \}$ and upper approximation $\overline{A} = \{ [\omega]_I \, | \, [\omega]_I \cap A \neq \emptyset \}$, respectively. The tuple $(\underline{A}, \overline{A})$ is called a rough set: $\underline{A}$ is the set of elements that are guaranteed to belong to $A$, whereas $\overline{A}$ is the set of elements that possibly belong to $A$. Fuzzy sets and rough sets capture complementary phenomena, namely graduality and granularity, and can be combined with each other, giving rise to fuzzy-rough sets or rough-fuzzy sets \citep{dubo_rf90}.

\section{A Unified View of Uncertainty}\label{sec:unified}
\begin{figure}
    \centering
    \includegraphics[width=.8\textwidth]{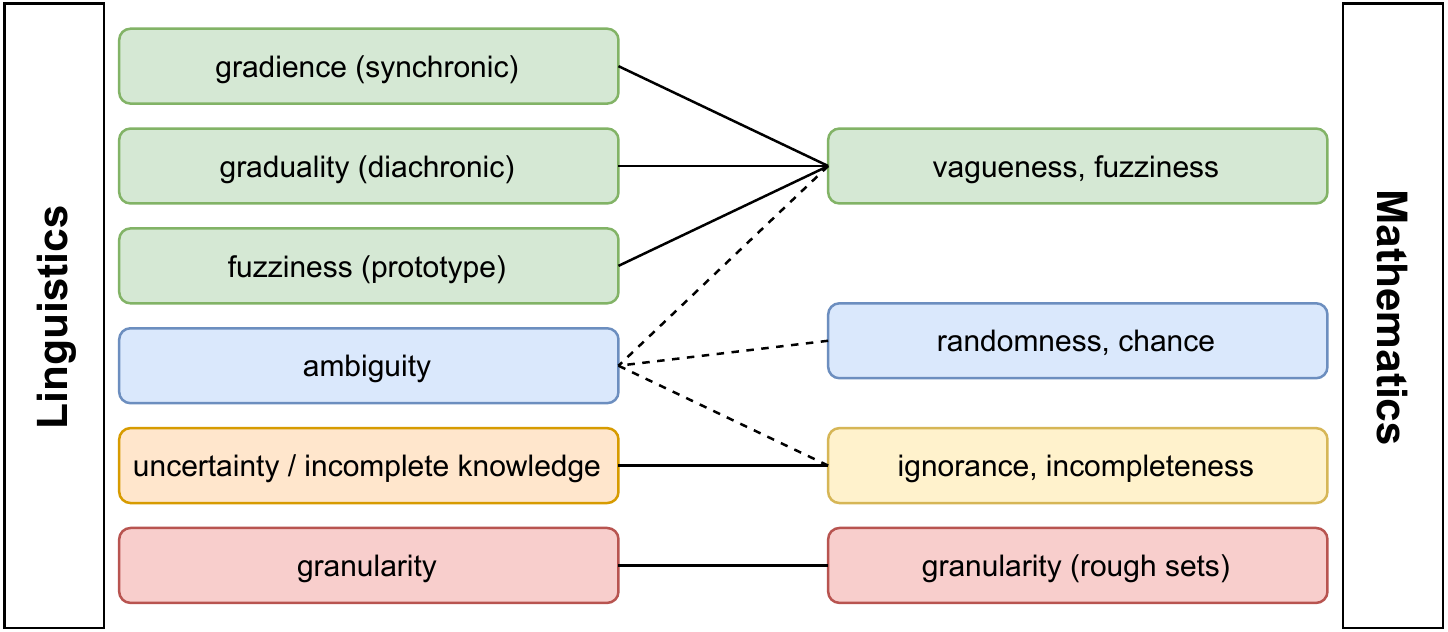}
    \caption{Mapping of the different notions of uncertainty in linguistics on the left and mathematics on the right side of the figure.}
    \label{fig:notions_of_uncertainty}
\end{figure}

In this section, we attempt to unify the views of uncertainty from linguistics and mathematics as outlined, respectively, in Sections~\ref{sec:linguistics-view} and \ref{sec:mathematical-view}. To this end, we establish a link between the key concepts and notions of uncertainty distinguished in the different disciplines and elaborate on how they relate to each other.
An overview of these links is given in Figure~\ref{fig:notions_of_uncertainty}, which provides the point of departure for our discussion.

\subsection{Fuzziness and Ambiguity}
On the mathematical side, properties such as non-sharp boundaries or gradual membership in a category are summarised under the notions of \textit{fuzziness} or \textit{vagueness} and formally modeled with the help of fuzzy sets.
While these notions largely agree with the view of vague phenomena in linguistics, the latter are further differentiated into sub-concepts such as \textit{gradience} and \textit{graduality}, thereby attempting to capture different types of fuzziness and their origin.

First of all, the view of linguistics differentiates between gradience (synchronic) and gradualness (diachronic).
Gradience, i.e., the absence of sharp boundaries for categories, is in turn differentiated into subsective and intersective gradience.
The former describes the phenomenon that an element can be a more or less good representative of a category, and thus can only be assigned to this category to a certain degree.
From the perspective of mathematics, this approach corresponds to that of fuzzy set theory, where elements of a set can also be more or less typical of a category, and where the notion of ``typicality'' \citep{krish94} or the related idea of a ``deformable prototype'' \citep{skala78} have been used for specifying membership functions. Obviously, the perspectives of linguistics and mathematics coincide in this view. Interestingly, very similar perspectives are also adopted in other disciplines, such as psychology \citep{hutt94}. 

Intersective gradience, on the other side, refers to the boundary between several categories. Such boundaries can be blurred, so that (the membership of) an element cannot be assigned to one of the categories uniquely. In other words, an element can fall into several categories at the same time, with different degrees of membership.
Mathematically, this is simply reflected by the fact that the element is (partially) covered by several fuzzy sets, i.e., the element lies in the intersection of the supports of these fuzzy sets\footnote{The support of a fuzzy set is the (non-fuzzy) set consisting of all elements with non-zero membership.}. Thus, even if subsective and intersective gradience adopt slightly different perspectives on a phenomenon, focusing, respectively, on the characteristics of a single category versus differences between two related categories, there is no real difference between them from a mathematical point of view. 

The phenomenon of graduality describes the development of categorical membership over a time dimension.
Over time, many micro-changes result in the emergence, change, and possibly disappearance of categories, so that the degrees of membership of all representatives of a category may change over time.
The temporal dimension is particularly important from the point of view of linguistics since keeping track of the \textit{development} of a language is only possible at this level. If one interpolates the many tiny steps in this development, it will result in a continuous shift of categorical boundaries.
From a mathematical point of view, this corresponds to the dynamic adaptation of fuzzy sets, defining membership degrees as a function of time \citep{lientz}.

Phenomena of ambiguity can be observed on different levels. 
On the functional level, i.e., the level of the word types or constructions, ambiguity results from the allowance of several readings regarding the role of words or groups of words\footnote{Interestingly, this ambiguity does not necessarily lead to ambiguity at the level of word semantics, where the message to the reader might still be the same.}.
That is, even if a ground truth construction can be assumed, it cannot be determined given the information at hand. Instead, based on that information, several interpretations appear plausible. Depending on the nature of the ambiguity and its causes, disambiguation through additional information (such as larger contexts of a construction in a text) may or may not be possible. If uncertainty cannot be reduced, the situation could be considered as a case of randomness from a mathematical point of view. Alternatively, one may argue that a unique ground truth in terms of a single interpretation does not exist, and instead declare the set of all plausible interpretations as the ground truth. Mathematically, these could be treated as an equivalence class or granule in the context of rough sets.


\subsection{Incompleteness and Lack of Knowledge}
\label{sec:inc}


As shown by our discussion so far, multiple dimensions of uncertainty may co-exist when annotating a corpus. In general, uncertainty spans over three dimensions: (1) The gradual plausibility of a specific 
tag, (2) the set of plausible tags (each of them being plausible to a certain degree), and (3) the epistemic uncertainty of the (human or machine) annotator. While (1) and (2) have essentially been covered by our discussion of (inter and intrasective) gradience and ambiguity, 
epistemic uncertainty in the sense of uncertainty due to incomplete knowledge and lacking annotation expertise is indeed another important aspect. 

Since human annotators usually act as non-native speakers, they need to \textit{learn} the language to be able to make accurate annotations.
Especially, when starting to annotate new material, they need to get familiar with the characteristics of the corpus.
Needless to say, human annotators should not be urged to make decisions too fast. Instead, they should be allowed and enabled to express and suitably represent their uncertainty.
At the same time, the human annotator must also remain vigilant and avoid pretending an unwarranted level of certainty that might be caused by the comparative fallacy problem.

Note that epistemic uncertainty is a kind of \emph{meta-uncertainty}, which does not directly refer to the underlying phenomena itself or any ground truth, but to the \emph{knowledge} about this ground truth, i.e., to the epistemic state of the annotator. Thus, epistemic uncertainty is \emph{reducible} uncertainty, i.e., uncertainty that can be reduced by gathering further information, whereas gradience, gradualness, and ambiguity are inherent properties of the language. Ideally, the human annotator would honestly document her uncertainty about the annotation itself according to her confidence.
This uncertainty can be expressed in different ways. In the \emph{set-based} approach, the annotator declares a subset of the entire set of tags as plausible candidates, merely distinguishing between tags that are definitely excluded and those that remain possible, at least tentatively. In the \emph{distributional} approach, the annotator provides scores or any other sort of rating for each tag, thereby distinguishing between plausibility and implausibility in a more fine-granular way. 

Obviously, the distributional approach provides more information than the set-based approach, which can be seen as a crude approximation of the former. Yet, for human annotators, it is often difficult to express their uncertainty or the extent to which a certain tag seems to be plausible, in terms of precise numbers. This is especially true in the presence of epistemic uncertainty, i.e., uncertainty about the quantification of uncertainty about the ground truth. 
Here, instead of working with precise numbers, it might be more appropriate to ask the annotator for qualitative judgments, e.g., to let her provide ratings on an ordinal plausibility scale. Even if this information is weaker than precise numerical degrees of plausibility or probability (for example, in the sense of being less discriminative between more plausible and less plausible options), it is likely to be more reliable.

Another important aspect concerns the question of whether the set of potential categories matches the closed or open world assumption. For example, tag sets for part-of-speech (POS) tagging are usually well defined, and typically there is no need for adding or removing new POS tags, nor does a single POS tag change its meaning over time. Still, certain parts of speech may have a completely different characteristic in the 13$^{th}$ century than the corresponding parts in the 16$^{th}$ century or even later. However, in the annotation process, it is crucial to annotate those parts equally to preserve consistency. Contrary to that, when annotating constructions, the tag sets need to be flexible enough to capture the emergence of new constructions, hence calling for an open world assumption. This has important implications on the mathematical modeling of the frame of discernment (cf.\ Section~\ref{sec:mathematical-view}). 

\begin{figure}[t]
    \centering
    \includegraphics[width=.9\linewidth]{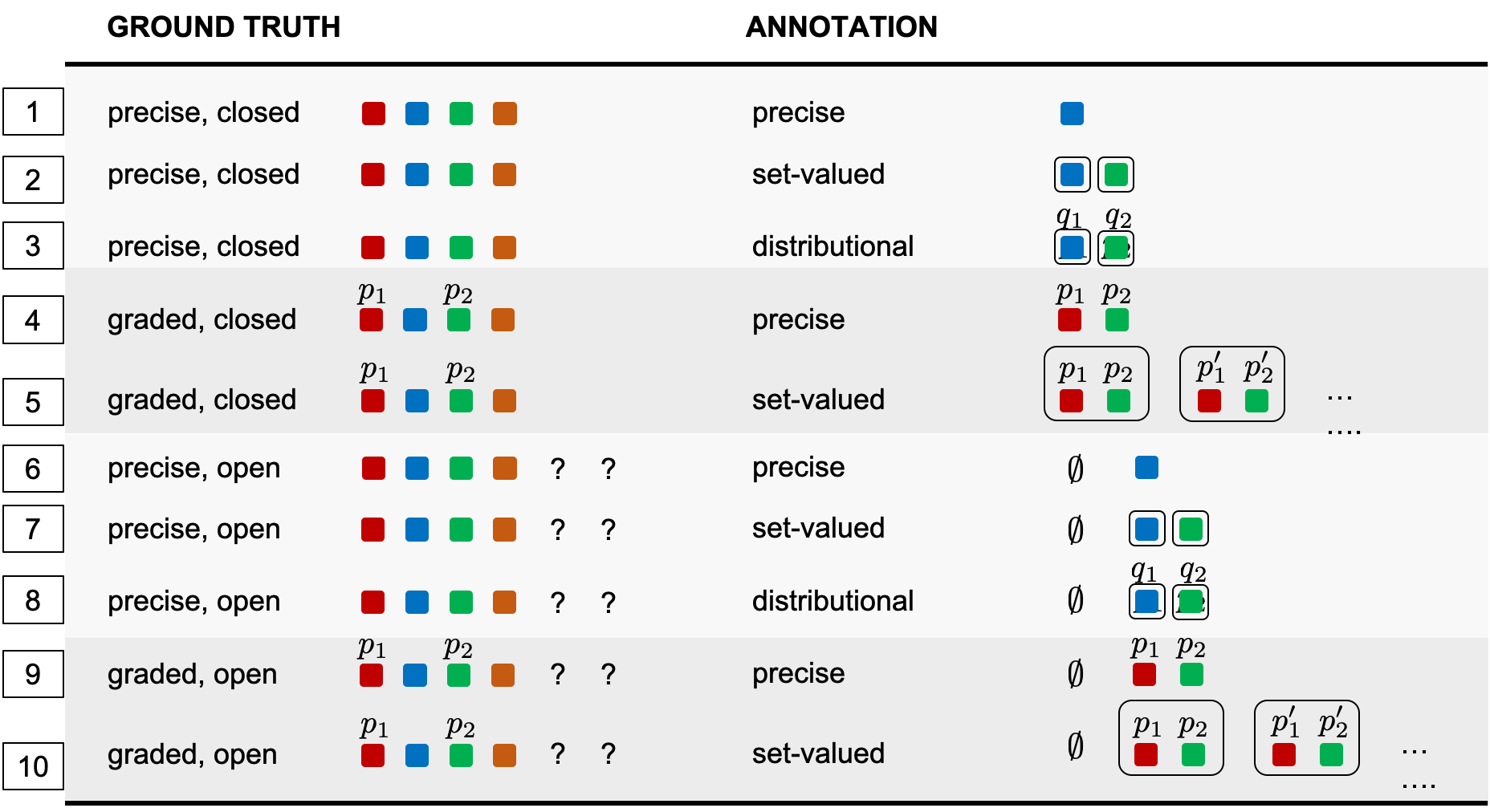}
    \caption{Different types of ground truth (precise or graded, open or closed world) and annotation (precise, set-valued, distributional), i.e., representation of knowledge about the ground truth. A distribution is indicated by membership degrees $p_i$ or $q_i$. In the case of a set-valued annotation, each element of the set (candidate annotation) is framed by a black line.}
    \label{fig:unc}
\end{figure}

\section{Practical Implications}\label{sec:practical-implications}

In this section, we discuss practical implications resulting from the previous, more theoretical considerations regarding different facets of uncertainty. The following discussions address the types of annotation we consider most relevant in practice, especially when dealing with historical corpora, as well as the gain of experience for both the human expert and machine annotator in the course of an annotation process.

\subsection{Different Types of Annotation}

Different ways of annotation are required for properly annotating (historical) texts, depending on what the properties of the ground truth are.
An overview of the different properties of the ground truth and corresponding annotations to represent these properties is given in Figure~\ref{fig:unc}.
On the left side of the figure, each line first describes the setting, i.e., whether the ground truth is precise or graded and whether we assume the tag set to be closed or open.
On the right side, this is then combined with a way of capturing the ground truth through different ways of annotation. 

Concerning POS tagging, current annotation practice is based on the assumption of a closed world with precisely definable categories: A predefined and fixed tag set is used.
Multiple taggings, i.e., that more than one part of speech applies to a token, is not intended (case 1).
In many cases, this indeed reflects daily practice in annotation projects.
For example, if we consider the sentence\\

\noindent
\begin{tabular}{r l}
& \textit{Dat is vredebrake}\\
Translation: & This is breach of peace \, ,
\end{tabular}\\

\noindent
exactly one POS tag can be assigned to each token:
The sentence is composed of a demonstrative pronoun Dat (DDS), a finite copular verb \textit{is} (VKFIN) and a noun appelativum \textit{vredebrake} (NA). However, the annotator might be unsure and consider two POS tags for a token possible – for example, both VKFIN and VAFIN (finite auxiliary verb) for the token\textit{ is }(case 2).
If necessary, he can also weight the possibilities that he cannot exclude, i.e., specify a distribution on the set of annotations.
Nevertheless, it is still assumed in this scenario (1 to 3) that only one tag is correct in the context of a closed world (predefined tag set).
If we assume gradient and therefore overlapping categories in the linguistic system, and if we take into account that a token can exhibit some properties of word category A and some properties of word category B, we operate with a graded ground truth.

The tag set, however, can nevertheless be predefined and closed, i.e., presuming that at least one tag, but possibly even more, can be assigned to each token.
Thus, the annotation can be either precise or set-valued. Let us recall the ambiguous example from Section 2 to illustrate this scenario:\\

\noindent
\begin{tabular}{r l}
     & \textit{Der folgende Abschnitt handelt vom in die Stadt fahren.}\\
     Verbatim: & The following section is about into the city drive.\\
     Translation: & The following section is about driving into the city.
\end{tabular}\\

The phrase [\textit{in die Stadt fahren}] can be interpreted as a verbal unit -- according to its origin -- or already as a unit with a nominal profile. In the first case, ``fahren'' is interpreted as a verb, the syntagma ``in die Stadt'' as an adverbial to this verb. In the nominal scenario ``fahren'' is already interpreted as a head noun, the prepositioned unit has a specifying entity (attribute). Both categorizations are correct, i.e., the ground truth is graded, the precise annotation also provides for the assignment of two tags (verb, noun).

Going beyond these 5 cases, annotation projects that are exploratory usually do not have a predefined and fixed tag set but rather develop a tag set on-the-fly.
This contradicts the closed world assumption and instead calls for an open world assumption (cases 6 to 10).
In this case, the tag set at a specific point in time might be incomplete in terms of coverage, i.e., there might be cases where the ground truth tag is not (yet) part of the tag set. Especially to cover a so far unknown range of phenomena, working with such an open tag set could be reasonable. Since in the underlying project the tag set for annotating complex constructions is created in parallel to the chronological exploration of the corpus texts, the inclusion of new tags is quite common, especially for more recent texts. It is only in the 15$^{th}$ century, for example, that a special prepositional construction indicating the punishment appears:\\

\noindent
\begin{tabular}{r p{9cm}}
     & \textit{Wie hijr theghen dede, dat is op die pene van eenre marken.} \citep[Article 487/6]{cleve} \\
     Translation: & Anyone who acts against, that's on the penalty of one mark.
\end{tabular}\\

In later texts, this punishment construction is used mainly adverbially and realized with different primary prepositions. A tag newly included in the course of the annotation covers these cases and thus takes into account a phenomenon that does not already occur at the beginning of the legal writing. Even in such cases, the annotator may be epistemically uncertain. Enabling him to use sets of tags in the annotation practice -- with or without weighting (set-valued or distributional) -- allows the human annotator to express this kind of uncertainty.
In addition to the assumption of an open world (= open tag set), gradient categories are also assumed in cases 9 and 10. Not only is the tagset extensible, but the categories are modeled as non-separable. A phenomenon at the language level can belong to both category A and category B. Here, an excerpt from the corpus, which has already been mentioned in Section~\ref{sec:linguistics-view}, can serve as an example:\\

\noindent
\begin{tabular}{r p{9cm}}
     &  \textit{We sik eruegu+odes vnderwint · oder an sprikt · \textbf{[[na]$_\text{APPR}$ [deme]$_\text{DDS}$ / [dat]$_\text{KOUS}$]$_\text{KOUS}$} it im vor delet is vor gherichte · Dat is en vredebrake} \citep{goslar} \\
     Translation: & Whoever takes possession of hereditary property or lays claim to it, \textbf{after it that (= after) }it has been deprived of it by judgment in court: this is a breach of the peace
\end{tabular}\\

On the one hand, the \textit{na deme dat} structure is both -- according to the original construction -- a prepositional phrase, and on the other hand, it exhibits properties of complex secondary subjunctions. Consequently, the ground truth in this case is graded. To tag the superordinate construction as such, a new construction tag is needed, since such structures do not appear in the earliest texts.

\subsection{Experience from the annotation practice:\\ Tagging ambiguities and uncertainties}
Experience in the scope of the outlined project has shown that the consideration and annotation of uncertainties, may it be epistemic uncertainty of the respective annotator or ambiguities in the material, are of great importance.
This concerns both the annotation by the human expert, to capture the whole reality and thus also salient examples of grammatical change, and the machine annotator, which is better able to handle complete and therefore imprecise data than data that is incomplete and additionally in some cases falsely annotated, just to meet some artificially created annotation convention.

Moreover, the closely coupled interaction between human and machine annotators is essential to be able to study a corpus with a sufficient coverage along the temporal and spatial dimensions that are needed to derive statistically robust statements.
However, key for this collaboration is the explicit exchange of information which needs to be understandable to the machine annotator and thus requires mathematical modeling.

\subsubsection{Human Expert Annotator}
The considerations presented strongly indicate an annotation with numerical values to record the interpretation of the data as accurately and adequately as possible. For the human annotator, however, this approach is not very practical, making the annotation task an almost impossible one to solve. Instead, it makes sense to work with an ordinal scale and to record different cases and degrees of uncertainty. Such an ordinal scale can be structured as follows, reference is made to the respective annotation tag(s) (POS / construction):

\begin{enumerate}
    \item definitely excluded (default)
    \item may apply, but unlikely
    \item not unplausible
    \item completely plausible
\end{enumerate}

This option of uncertainty annotation turns out to be much more user-friendly and has also been chosen for the underlying project. Both multiple annotations and annotations using a predefined uncertainty tag set were recorded with appropriate tool support, i.e., the annotation of uncertainties is systematic and can be traced across all texts. 

If the annotator is allowed to assign such plausibilities to different tags, annotations like (A/3, B/2) are produced, i. e., tag A is completely plausible and B at least not implausible. However, it is not possible to tell from such an annotation whether the annotator assumes a precise or gradual ground truth, whether the annotation rather encodes a distribution over the tags or a distribution over the distributions over the tags. For the machine learner, it is therefore not clear how to interpret such training information. Even if the annotator cannot be expected to specify explicitly uncertain knowledge about a gradual ground truth, it seems to make sense to ask the human annotator at least one more bit of information: namely, whether he considers the ground truth to be precise or gradual, i.e., whether A and B are exclusive in the example or whether both can apply simultaneously.

What are the advantages of such a procedure for manual annotation (in larger annotation teams)? Instead of establishing conventions to ensure consistency and enforcing certain interpretations, it becomes possible to annotate different interpretations and to specify degrees of uncertainty for these annotations. Allowing for multiple (uncertain) interpretations prevents agreements on tagging solutions which only reflect the partial truth, since multiple interpretations can indeed be correct. Annotation guidelines based on agreement on one single POS and/or construction annotation distort linguistic reality and thus block the most immediate and appropriate access to phenomena of grammatical change. In this manner, for example, bridging and switching contexts \citep{hei02,heinar10,diewald2009konstruktionen} can be taken into account in the context of the grammaticalization of linguistic units. Constructional variability and phenomena of linguistic vagueness are also more likely to be taken into consideration. Also with regard to the machine annotator, which is trained based on human expert's annotation (as smaller training corpus), this approach using an ordinal scale (predefined uncertainty tag set) makes sense: It is considerably better to imprecisely annotate the training corpus, i.e., to make uncertainties explicit, instead of deliberately feeding wrong/incomplete annotations into a machine annotator.

The following is less a disadvantage than a fundamental problem in a highly interpretative research context such as tagging different types of uncertainty: Sometimes the decision whether a text passage is ambiguous and the ground truth is a graded one or the annotator lacks the knowledge to understand the text is very clear. In some places, however, this question can be answered less clearly. Uncertainties based on structural ambiguity and those resulting from a lack of knowledge cannot always be clearly separated by human expert annotators. Moreover, the human annotator tends far more often to blame missing knowledge on his part -- an experience of the project. Sentences/structures in the text are reminiscent of familiar ambiguities and developmental scenarios. It may be obvious to tag these text passages as ambiguous, but individual word material causes difficulties. The annotator is unsure which uncertainty tag to choose. Also, the increasing experience with annotation (as well as (canonical) knowledge of universal grammaticalization paths) influences uncertainty decisions that should not be underestimated. Human annotation moves between increasing routinization, the knowledge of analogy in the corpus, and existing bridges to known linguistic research results.

\subsubsection{Machine Annotator}

When dealing with a large corpus which is inevitable if a robust statistical analysis is to be carried out, this corpus needs to be annotated thoroughly.
Since such a corpus can easily comprise millions or even tens of millions of tokens, the sole annotation process becomes a time-consuming task which yet only represents the basis for the actual task of interest, i.e., the analysis.
In turn, the size of the corpus that can be considered within a (fixed-term) project is quite limited if only relying on human experts to annotate the respective corpus.

Fortunately, the annotation process can be sped up considerably employing a machine annotator.
Building mathematical models based on parts of the corpus that are already annotated by the human annotator, the annotations for the remaining corpus can be automatically completed.
As the parts annotated are considered examples of how the human annotator \textit{would} label the entire corpus and the machine annotator will try to imitate this behavior, false annotations might be tolerated if they are not overly frequent but will still complicate the induction of an annotation model, e.g., a part-of-speech tagger.
However, if uncertainties or ambiguities are ignored during the annotation of the data to induce the annotation model from, it will also enforce the model to \textit{learn} to imitate the false annotation behavior.
This problem may be due to the annotator exhibiting a certain bias (consciously or subconsciously) towards a certain annotation or because there is simply not a single ground truth.
If the latter is the case, it needs to be taken into account when defining the model that it is possible that potentially there is more than a single ground truth and, in such cases, examples also need to be shown to the machine annotator to be able to cope with such ambiguities.

Consequently, the effectiveness of a machine annotator behaves proportionally to the quality of the data annotated by the human annotator.
Analogously, low-quality human annotations will also yield inaccurate or biased annotation models.

To make uncertainties and ambiguities recognizable and understandable to the machine annotator, suitable mathematical formalisms as described in Section~\ref{sec:mathematical-view} are necessary to process them consistently in the induction process of annotations models.
Furthermore, it is equally important that the machine annotator is not only able to understand uncertainties and ambiguities to build hypotheses about the annotation task, but also to communicate its (model) uncertainty \citep{heid2020reliable}, which is different than the epistemic uncertainty of the human expert, and identified ambiguities to the human annotator.
Thereby, the machine and the human annotator can collaborate on the annotation task at an equal level.

Another advantage of making uncertainties and ambiguities explicit is that it allows automated analysis of the data for statistical evaluation.
This includes significant passages where grammatical change manifests itself in the form of ambiguities as well as to systematically track transitions of elements between categories and thereby keep track of the development of a language.

\section{Conclusion}\label{sec:conclusion}
For corpus linguistic annotation projects, it is crucial to deal with and acknowledge the existence of uncertainty which in some cases is reflected in terms of ambiguities.
To increase annotation quality, it is important to make uncertainty explicit.
More precisely, all valid interpretations should be captured in case of ambiguity instead of setting up conventions defining which interpretation to choose. Another important aspect to consider is the uncertainty of the annotator him- or herself.
On one hand, annotations that are made with high uncertainty are error-prone and thus may lead to wrong conclusions in a subsequent analysis of the annotated data. On the other hand, when the annotation process is to be supported by a machine annotator, the explicit statement of uncertainty is necessary to communicate this uncertainty to the respective algorithm. Obviously, a wrong annotation with high uncertainty will not mislead the machine annotator as a wrong annotation with (implicit) high certainty.

However, the correct formalization of uncertainty requires a close collaboration of various disciplines.
Aligning the notions of uncertainty developed in the respective fields in the first place is a crucial requirement.
In this paper we made a first step towards such an alignment, unifying the notions of uncertainty from linguistics and mathematics, outlining practical implications, and raising additional questions that are left for future work.
Combining the expertise of the different fields was a key enabler for this alignment and is inevitable for upcoming works.

In future work, we want to provide empirical evidence that taking uncertainties into account is not only beneficial in theory and in the considered examples but also improves the annotation quality on a broader scale.
Additionally, we want to extend current approaches for automating the annotation employing machine annotators which are capable of both, learning from texts containing annotated uncertainties as well as expressing the uncertainty of the machine annotator.
The latter increases the efficiency of manually revising the automatically generated annotations since parts with high uncertainty (and thus more likely containing errors) are explicitly stated and no longer need to be detected by the human expert him- or herself.

\bibliographystyle{plainnat}
\bibliography{literature}

\begin{thebibliography}{53}
\providecommand{\natexlab}[1]{#1}
\providecommand{\url}[1]{\texttt{#1}}
\expandafter\ifx\csname urlstyle\endcsname\relax
  \providecommand{\doi}[1]{doi: #1}\else
  \providecommand{\doi}{doi: \begingroup \urlstyle{rm}\Url}\fi

\bibitem[Aarts(2007)]{aar07}
Bas Aarts.
\newblock \emph{Syntactic Gradience. The Nature of Grammatical Indeterminacy}.
\newblock Cambridge University Press, New York, 2007.

\bibitem[Bley-Vroman(1983)]{ble83}
Richard Bley-Vroman.
\newblock The comparative fallacy in interlanguage studies: the case of
  systematicity.
\newblock In \emph{Language Learning}, volume~33, pages 1--17. 1983.

\bibitem[Buckley(2006)]{buckley}
J.J.\ Buckley.
\newblock \emph{Fuzzy Probability and Statistics}.
\newblock Springer, 2006.

\bibitem[Bybee(2010)]{byb10}
Joan~L. Bybee.
\newblock \emph{Language, Usage and Cognition}.
\newblock Cambridge University Press, New York, 2010.

\bibitem[Bybee(2011)]{byb11}
Joan~L. Bybee.
\newblock Usage-based theory and grammaticalization.
\newblock In Heiko Narrog and Bernd Heine, editors, \emph{The Oxford Handbook
  of Grammaticalization}, pages 60--78. Oxford University Press, Oxford/New
  York, 2011.

\bibitem[Croft(2001)]{cro01}
William~A. Croft.
\newblock \emph{Radical Construction Grammar: Syntactic Theory in Typological
  Perspective}.
\newblock Oxford University Press, New York, 2001.

\bibitem[Deng(2014)]{deng_ge14}
Y.\ Deng.
\newblock Generalized evidence theory.
\newblock \emph{CoRR}, abs/404.4801.v1, 2014.
\newblock URL \url{http://arxiv.org/abs/404.4801}.

\bibitem[Denison(2017)]{den17}
David Denison.
\newblock Ambiguity and vagueness in historical change.
\newblock In M.~Hundt, S.~Molling, and S.~E. Pfenniger, editors, \emph{The
  Changing English Language. Psycholinguistic Perspectives}, pages 292--318.
  Cambridge University Press, New York, 2017.

\bibitem[Diewald(2009)]{diewald2009konstruktionen}
Gabriele Diewald.
\newblock Konstruktionen und paradigmen.
\newblock \emph{Zeitschrift f{\"u}r germanistische Linguistik}, 37\penalty0
  (3):\penalty0 445--468, 2009.

\bibitem[Dipper(2015)]{dip15}
Stefanie Dipper.
\newblock {Annotierte Korpora für die Historische Syntaxforschung:
  Anwendungsbeispiele anhand des Referenzkorpus Mittelniederdeutsch}.
\newblock \emph{Zeitschrift für Germanistische Linguistik}, 43(3):\penalty0
  516--563, 2015.

\bibitem[Dipper et~al.(2013)Dipper, Donhauser, Klein, Linde, M{\"u}ller, and
  Wegera]{dipdonkle13}
Stefanie Dipper, Karin Donhauser, Thomas Klein, Sonja Linde, Stefan M{\"u}ller,
  and Klaus-Peter Wegera.
\newblock {HiTS: ein Tagset f{\"{u}}r historische Sprachstufen des Deutschen}.
\newblock \emph{Journal for Language Technology and Computational Linguistics},
  28:\penalty0 85--137, 2013.

\bibitem[Dubois(2006)]{dubo_pt06}
D.\ Dubois.
\newblock Possibility theory and statistical reasoning.
\newblock \emph{Computational Statistics and Data Analysis}, 51\penalty0
  (1):\penalty0 47--69, 2006.

\bibitem[Dubois and Prade(1988)]{dubo_pt}
D.~Dubois and H.~Prade.
\newblock \emph{Possibility Theory}.
\newblock Plenum Press, 1988.

\bibitem[Dubois and Prade(1990)]{dubo_rf90}
D.~Dubois and H.~Prade.
\newblock Rough fuzzy sets and fuzzy rough sets.
\newblock \emph{Int.\ J.\ General Systems}, 17:\penalty0 191--209, 1990.

\bibitem[Dubois et~al.(1996)Dubois, Prade, and Smets]{dubo_rp96}
D.~Dubois, H.~Prade, and P.~Smets.
\newblock Representing partial ignorance.
\newblock \emph{IEEE Transactions on Systems, Man and Cybernetics, Series A},
  26\penalty0 (3):\penalty0 361--377, 1996.

\bibitem[Eckart~de Castilho et~al.(2016)Eckart~de Castilho,
  M{\'u}jdricza-Maydt, Yimam, Hartmann, Gurevych, Frank, and
  Biemann]{eckart-de-castilho-etal-2016-web}
Richard Eckart~de Castilho, {\'E}va M{\'u}jdricza-Maydt, Seid~Muhie Yimam,
  Silvana Hartmann, Iryna Gurevych, Anette Frank, and Chris Biemann.
\newblock A web-based tool for the integrated annotation of semantic and
  syntactic structures.
\newblock In \emph{Proceedings of the Workshop on Language Technology Resources
  and Tools for Digital Humanities ({LT}4{DH})}, pages 76--84, Osaka, Japan,
  December 2016. The COLING 2016 Organizing Committee.
\newblock URL \url{https://www.aclweb.org/anthology/W16-4011}.

\bibitem[Flink(1991)]{cleve}
Klaus Flink.
\newblock Das stadtrecht von cleve (klever archiv 11).
\newblock Kleve: Selbstverlag des Stadtarchivs Kleve, 1991.
\newblock 1424 urban law of Cleve.

\bibitem[Grice(1975)]{grice3p}
Herbert Grice.
\newblock P.(1975). logic and conversation.
\newblock \emph{Syntax and semantics}, 3:\penalty0 43--58, 1975.

\bibitem[Hacking(1975)]{hack_te}
I.\ Hacking.
\newblock \emph{The emergence of probability: {A} philosophical study of early
  ideas about probability, induction and statistical inference}.
\newblock Cambridge University Press, New York, 1975.

\bibitem[Hajek(1998)]{haje_mf}
P.~Hajek.
\newblock \emph{Metamathematics of Fuzzy Logic}.
\newblock Kluwer, Dordrecht, 1998.

\bibitem[Heid et~al.(2020)Heid, Wever, and H{\"u}llermeier]{heid2020reliable}
Stefan Heid, Marcel Wever, and Eyke H{\"u}llermeier.
\newblock Reliable part-of-speech tagging of historical corpora through
  set-valued prediction.
\newblock \emph{arXiv preprint arXiv:2008.01377}, 2020.

\bibitem[Heine(2002)]{hei02}
Bernd Heine.
\newblock On the role of context in grammaticalization.
\newblock In Ilse Wischer and Gabriele Diewald, editors, \emph{New reflections
  on grammaticalization}, pages 83--101. John Benjamins, Amsterdam, 2002.

\bibitem[Heine and Narrog(2010)]{heinar10}
Bernd Heine and Heiko Narrog.
\newblock Grammaticalization and linguistic analysis.
\newblock In Bernd Heine and Heiko Narrog, editors, \emph{The Oxford Handbook
  of Linguistic Analysis}, pages 401--423. Oxford University Press, New York,
  2010.

\bibitem[Huttenlocher and Hedges(1994)]{hutt94}
J.\ Huttenlocher and L.V.\ Hedges.
\newblock Combining graded categories: {M}embership and typicality.
\newblock \emph{Psychological Review}, 101\penalty0 (1):\penalty0 157--165,
  1994.

\bibitem[Kübler and Zinsmeister(2015)]{kueblerZinsmeister2015}
Sandra Kübler and Heike Zinsmeister.
\newblock \emph{Corpus Linguistics and Linguistically Annotated Corpora}.
\newblock London: Bloomsbury Publishing, 2015.

\bibitem[Klie et~al.(2018)Klie, Bugert, Boullosa, de~Castilho, and
  Gurevych]{tubiblio106270}
Jan-Christoph Klie, Michael Bugert, Beto Boullosa, Richard~Eckart de~Castilho,
  and Iryna Gurevych.
\newblock The inception platform: Machine-assisted and knowledge-oriented
  interactive annotation.
\newblock In \emph{Proceedings of the 27th International Conference on
  Computational Linguistics: System Demonstrations}, pages 5--9. Association
  for Computational Linguistics, Juni 2018.
\newblock URL \url{http://tubiblio.ulb.tu-darmstadt.de/106270/}.

\bibitem[Krishnapuram(1994)]{krish94}
R.~Krishnapuram.
\newblock Generation of membership functions via possibilistic clustering.
\newblock In \emph{Proc.\ IEEE 3rd International Fuzzy Systems Conference},
  1994.

\bibitem[Kruse et~al.(1991)Kruse, Schwecke, and Heinsohn]{krus_ua}
R.~Kruse, E.~Schwecke, and J.~Heinsohn.
\newblock \emph{Uncertainty and Vagueness in Knowledge Based Systems}.
\newblock Springer-Verlag, 1991.

\bibitem[Lakoff(1987)]{lak87}
George Lakoff.
\newblock Cognitive models and prototype theory.
\newblock In Ulric Neisser, editor, \emph{Concepts and conceptual development},
  pages 63--100. Cambridge University Press, Cambridge, 1987.

\bibitem[Langacker(1987)]{lan87}
Ronald~W. Langacker.
\newblock \emph{Foundations of Cognitive Grammar (I). Theoretical
  Prerequisites}.
\newblock Stanford University Press, Stanford, 1987.

\bibitem[Lehmberg(2013)]{goslar}
Maik Lehmberg.
\newblock Der goslarer ratskodex – das stadtrecht um 1350: Edition,
  Übersetzung und begleitende beiträge.
\newblock Bielefeld: Verlag für Regionalgeschichte, 2013.
\newblock 1350 urban law of Goslar.

\bibitem[Lientz(1972)]{lientz}
Bennet~P.\ Lientz.
\newblock On time dependent fuzzy sets.
\newblock \emph{Information Sciences}, 4\penalty0 (3--4):\penalty0 367--376,
  1972.

\bibitem[Maas(2010)]{maas2010}
Utz Maas.
\newblock Einleitung. / literat und orat. grundbegriffe der analyse
  geschriebener und gesprochener sprache.
\newblock \emph{Grazer linguistische Studien}, 73:\penalty0 5--150, 2010.

\bibitem[Matheron(1975)]{math_rs}
G.~Matheron.
\newblock \emph{Random Sets and Integral Geometry}.
\newblock John Wiley and Sons, 1975.

\bibitem[Merten(2018)]{mer18}
Marie-Luis Merten.
\newblock \emph{{Literater Sprachausbau kognitiv-funktional.
  Funktionswort-Konstruktionen in der historischen Rechtsschriftlichkeit}}.
\newblock de Gruyter, Berlin/Boston, 2018.

\bibitem[Merten and Tophinke(2019)]{mertentophinke2019}
Marie-Luis Merten and Doris Tophinke.
\newblock Interaktive analyse historischen grammatikwandels.
  konstruktionsgrammatik trifft auf machine learning.
\newblock \emph{Jahrbuch für Germanistische Sprachgeschichte}, 10:\penalty0
  303--323, 2019.

\bibitem[Nguyen(1978)]{nguy_or78}
H.T. Nguyen.
\newblock On random sets and belief functions.
\newblock \emph{Journal of Mathematical Analysis and Applications},
  65:\penalty0 531--542, 1978.

\bibitem[Pawlak(1982)]{pawl_rs82}
Z.~Pawlak.
\newblock Rough sets.
\newblock \emph{International Journal of Computer and Information Sciences},
  11:\penalty0 341--356, 1982.

\bibitem[Russell(1923)]{russell}
B.~Russell.
\newblock Vagueness.
\newblock \emph{The Australasian Journal of Psychology and Philosophy},
  1:\penalty0 84--92, 1923.

\bibitem[Schmid(2010)]{sch10}
Hans-Jörg Schmid.
\newblock \emph{{Does frequency in text instatiate entrenchment in the
  cognitive system?}}
\newblock de Gruyter, Berlin, 2010.

\bibitem[Seemann et~al.(2017)Seemann, Merten, Geierhos, Tophinke, and
  H{\"u}llermeier]{seemann-etal-2017-annotation}
Nina Seemann, Marie-Luis Merten, Michaela Geierhos, Doris Tophinke, and Eyke
  H{\"u}llermeier.
\newblock Annotation challenges for reconstructing the structural elaboration
  of middle low {G}erman.
\newblock In \emph{Proceedings of the Joint {SIGHUM} Workshop on Computational
  Linguistics for Cultural Heritage, Social Sciences, Humanities and
  Literature}, pages 40--45, Vancouver, Canada, 2017. Association for
  Computational Linguistics.

\bibitem[Shafer(1976)]{shaf_am}
G.~Shafer.
\newblock \emph{A Mathematical Theory of Evidence}.
\newblock Princeton University Press, 1976.

\bibitem[Shilkret(1971)]{shil_mm71}
N.\ Shilkret.
\newblock Maxitive measure and integration.
\newblock \emph{Nederl.\ Akad.\ Wetensch.\ Proc.\ Ser.\ A 74 = Indag.\ Math.},
  33:\penalty0 109--116, 1971.

\bibitem[Skala(1978)]{skala78}
H.J.\ Skala.
\newblock On many-valued logics, fuzzy sets, fuzzy logics and their
  applications.
\newblock \emph{Fuzzy Sets and Systems}, 1\penalty0 (2):\penalty0 129--149,
  1978.

\bibitem[Smets and Kennes(1994)]{smet_tt94}
P.~Smets and R.~Kennes.
\newblock The transferable belief model.
\newblock \emph{Artificial Intelligence}, 66:\penalty0 191--234, 1994.

\bibitem[Taylor(2003)]{tay03}
John~R. Taylor.
\newblock \emph{Linguistic Categorization}.
\newblock Oxford University Press, New York, 2003.

\bibitem[Tophinke(2012)]{tophinke2012}
Doris Tophinke.
\newblock Syntaktischer ausbau im mittelniederdeutschen.
  theoretisch-methodische {\"{u}}berlegungen und kursorische analysen.
\newblock \emph{Niederdeutsches Wort}, 52:\penalty0 19--46, 2012.

\bibitem[Traugott and Trousdale(2010)]{tratro10}
Elizabeth~C. Traugott and Graeme Trousdale.
\newblock Gradience, gradualness and grammaticalization: How do they intersect?
\newblock In Elizabeth~C. Traugott and Graeme Trousdale, editors,
  \emph{Gradience, Gradualness and Grammaticalization}, pages 19--44. John
  Benjamins, Amsterdam, 2010.

\bibitem[Traugott and Trousdale(2013)]{tratro13}
Elizabeth~C. Traugott and Graeme Trousdale.
\newblock \emph{Constructionalization and constructional changes}.
\newblock Oxford University Press, Oxford, 2013.

\bibitem[Trousdale(2012)]{trousdale2012}
Graeme Trousdale.
\newblock Grammaticalization, lexicalization and constructionalization from a
  cognitive-pragmatic perspective.
\newblock In Hans-Jörg Schmid, editor, \emph{Cognitive Pragmatics}, pages
  533--558. Berlin, New York: de Gruyter, 2012.

\bibitem[Trousdale(2013)]{trousdale2013gradualness}
Graeme Trousdale.
\newblock Gradualness in language change.
\newblock \emph{Synchrony and Diachrony: A dynamic interface}, 133:\penalty0
  27, 2013.

\bibitem[Walley(1991)]{wall_sr}
P.~Walley.
\newblock \emph{Statistical Reasoning with Imprecise Probabilities}.
\newblock Chapman and Hall, 1991.

\bibitem[Zadeh(1965)]{zade_fs65}
L.A. Zadeh.
\newblock Fuzzy sets.
\newblock \emph{Information and Control}, 8\penalty0 (3):\penalty0 338--353,
  1965.

\end{thebibliography}

\end{document}